\documentclass[runningheads]{llncs}

\usepackage{amsmath}
\usepackage[T1]{fontenc}
\usepackage{graphicx}
\usepackage{booktabs}
\usepackage{subcaption}
\usepackage{soul}
\usepackage{xcolor}
\usepackage{comment}

\usepackage{cite}

\begin{document}
%
\title{Engagement Measurement Based on Facial Landmarks and Spatial-Temporal Graph Convolutional Networks}

\author{Ali Abedi\inst{1} \and
Shehroz S. Khan\inst{1,2}}
\authorrunning{Ali Abedi and Shehroz S. Khan}
%
\institute{KITE Research Institute, University Health Network, Toronto, Canada \and
Institute of Biomedical Engineering, University of Toronto, Toronto, Canada
\\
\email{\{ali.abedi,shehroz.khan\}@uhn.ca}}
\maketitle              
\begin{abstract}
Engagement in virtual learning is crucial for a variety of factors including student satisfaction, performance, and compliance with learning programs, but measuring it is a challenging task. There is therefore considerable interest in utilizing artificial intelligence and affective computing to measure engagement in natural settings as well as on a large scale. This paper introduces a novel, privacy-preserving method for engagement measurement from videos. It uses facial landmarks, which carry no personally identifiable information, extracted from videos via the MediaPipe deep learning solution. The extracted facial landmarks are fed to Spatial-Temporal Graph Convolutional Networks (ST-GCNs) to output the engagement level of the student in the video. To integrate the ordinal nature of the engagement variable into the training process, ST-GCNs undergo training in a novel ordinal learning framework based on transfer learning. Experimental results on two video student engagement measurement datasets show the superiority of the proposed method compared to previous methods with improved state-of-the-art on the EngageNet dataset with a 3.1\% improvement in four-class engagement level classification accuracy and on the Online Student Engagement dataset with a 1.5\% improvement in binary engagement classification accuracy. Gradient-weighted Class Activation Mapping (Grad-CAM) was applied to the developed ST-GCNs to interpret the engagement measurements obtained by the proposed method in both the spatial and temporal domains. The relatively lightweight and fast ST-GCN and its integration with the real-time MediaPipe make the proposed approach capable of being deployed on virtual learning platforms and measuring engagement in real-time.

\end{abstract}

\vspace{-2mm}
\section{Introduction}
\vspace{-2mm}
Engagement is key in education, blending students' attention and interest within a learning context \cite{10243609}. It not only stems from existing interests but also fosters new ones through sustained attention \cite{hidi2006four}, essential for content comprehension and engagement development \cite{10243609}. However, measuring and upholding engagement is challenging, and demands significant effort from educators. The advent of remote sensing, Artificial Intelligence (AI), and affective computing offers new avenues for accurately measuring engagement across various learning environments, including virtual learning platforms. Technologies such as facial expression recognition \cite{liu2022graph} and eye gaze tracking \cite{wood2015rendering} enable more precise monitoring and enhancement of student engagement. This paper explores the development of AI algorithms for automated engagement measurement in virtual learning, taking into account the definition of engagement in educational psychology and its measurement methodologies.

\vspace{-.5mm}
Fredricks et al. \cite{fredricks2004school} defined engagement through affective, behavioral, and cognitive components. Affective engagement relates to students' emotions and attitudes towards their tasks, including interest or feelings during a learning class \cite{d2012dynamics}. Behavioral engagement refers to active participation, such as focusing on a computer screen or not being distracted by a phone \cite{ocumpaugh2015baker}. Cognitive engagement deals with a student's dedication to learning and tackling challenges, affecting outcomes such as recall and understanding \cite{10243609}. Booth et al. \cite{10243609} identified that engagement is measured through various signals such as facial expressions, eye movements, posture, heart rate, brain activity, audio, and digital interactions. Cameras, prevalent in devices for online learning, make video a key data modality for AI-based engagement measurement \cite{karimah2022automatic,dewan2019engagement}. Therefore, AI techniques predominantly focus on video to measure student engagement \cite{karimah2022automatic,dewan2019engagement,khan2022inconsistencies}.

AI-driven engagement measurement methods are divided into end-to-end and feature-based approaches. While feature-based methods generally outperform end-to-end approaches, which process raw videos, they require extensive identification of effective features through trial and error \cite{karimah2022automatic,dewan2019engagement,khan2022inconsistencies}. Extracting multi-modal features from videos with multiple computer vision algorithms or neural networks \cite{cai2023marlin,engagenet}, and their subsequent analysis by deep neural networks to measure engagement, render these methods computationally demanding \cite{abedi2023affect,karimah2022automatic}. Such computational requirements limit their use on local devices and necessitate the transfer of privacy-sensitive video data to cloud servers for engagement measurement. In contrast, extracting low-dimensional landmark information, such as facial and hand landmarks, not only provides more compact data but also captures essential geometric features for affect and behavior analysis, including engagement, without personal identifiers \cite{lugaresi2019mediapipe,wei2023geometric,10144079}. Previous feature-based engagement measurement approaches extracted features such as head pose, facial Action Units (AUs), iris and gaze features, and affect and emotion features, which are indicators of the behavioral and affective components of engagement \cite{karimah2022automatic, dewan2019engagement, khan2022inconsistencies}. The literature has demonstrated the success of facial landmarks in capturing these aforementioned features \cite{malek2021head, jacob2021facial, toisoul2021estimation, grishchenko2020attention, liu2022graph}. A model such as Spatial-Temporal Graph Convolutional Networks (ST-GCNs) \cite{yan2018spatial}, capable of analyzing spatial-temporal facial landmarks and learning these aforementioned features inherently, can reduce the need for raw facial videos. An approach based on facial landmarks prioritizes privacy and reduces computational demands, making it more practical for real-time engagement measurement.

In this paper, an alternative course away from conventional end-to-end and feature-based approaches is charted, and a novel engagement measurement technique using facial landmarks extracted from videos is presented. The proposed method is characterized by its privacy-preserving nature and computational efficiency. This work makes the following contributions:
\begin{itemize}
  \item This marks the first instance in video-based engagement measurement \cite{karimah2022automatic,dewan2019engagement,ma2021automatic,copur2022engagement,abedi2023affect,abedi2023detecting,fwa2022fine} where facial landmarks, as the single data modality extracted from videos, are analyzed through ST-GCNs \cite{yan2018spatial} to infer the engagement level in the video;
  \item To integrate the ordinal nature of the engagement variable into the training process, ST-GCNs undergo training in a novel ordinal learning framework utilizing transfer learning;
  \item Extensive experiments conducted on two video-based engagement measurement datasets demonstrate the superiority of the proposed method over previous methods, achieving an improved state-of-the-art in engagement level classification accuracy. For explainability, Gradient-weighted Class Activation Mapping (Grad-CAM) is applied to the developed ST-GCNs to interpret the engagement measurements obtained by the proposed method in both spatial and temporal domains.
\end{itemize}


\vspace{-2mm}
\section{Related Work}
\label{sec:related_work}
\vspace{-2mm}
The literature review in this paper serves two main purposes: discussing past video-based engagement measurement techniques and examining the use of graph convolutional networks for facial expression and affect analysis.

\vspace{-2mm}
\subsection{Engagement Measurement}
\label{engagement_measurement}
Karimah et al. \cite{karimah2022automatic} conducted a systematic review on measuring student engagement in virtual learning environments, revealing a focus on affective and behavioral components of engagement. Most methods utilize datasets annotated by external observers \cite{khan2022inconsistencies} to train both feature-based and end-to-end models. In the following, some of the relevant feature-based and end-to-end works on video-based engagement measurement are discussed.

In the domain of end-to-end engagement measurement techniques, deep neural networks analyze consecutive raw video frames to output the engagement level of the student in the video. These methods do not employ the extraction of handcrafted features from the videos; instead, the network is adept at autonomously learning to extract the most useful features directly from the videos, utilizing consecutive convolutional layers. The deep neural networks implemented in these end-to-end approaches include networks capable of video analysis, such as 3D Convolutional Neural Networks (3D CNNs) and Video Transformers \cite{gupta2016daisee,abedi2021improving,ai2022class}, as well as combinations of 2D CNNs with sequential neural networks such as Long Short-Term Memory (LSTM) and Temporal Convolutional Network (TCN) \cite{gupta2016daisee,abedi2021improving,liao2021deep,selim2022students}.

The process of measuring engagement through feature-based techniques involves two stages. Initially, behavioral and affective features are extracted from video frames, relying on either domain-specific knowledge or pre-trained models for facial embedding extraction. OpenFace \cite{baltrusaitis2018openface} is notably effective for its comprehensive feature extraction capabilities, including AUs, eye movements, gaze direction, and head positioning, and is widely applied in engagement measurement \cite{abedi2023affect,abedi2023bag,thomas2018predicting,engagenet}. Examples of facial embedding include the Masked Autoencoder for facial video Representation LearnINg (MARLIN) \cite{cai2023marlin}, utilized by Singh et al. \cite{engagenet}, and the Emotion Face Alignment Network (EmoFAN) \cite{toisoul2021estimation}, used by Abedi et al. \cite{abedi2023affect}. Subsequently, to analyze these extracted features and infer engagement, various machine learning and deep learning models are employed, such as Bag-of-Words (BoW) \cite{abedi2023bag}, Recurrent Neural Network (RNN) variations \cite{abedi2023detecting}, Temporal Convolutional Networks (TCNs) \cite{abedi2023affect,thomas2018predicting}, Transformers \cite{engagenet,vedernikov2024tcct}, and ensemble models \cite{tian2023predicting}.


\vspace{-2mm}
\subsection{Graph-based Facial Affect and Expression Analysis}
\label{graph_analysis}
\vspace{-2mm}
Liu et al. \cite{liu2022graph} conducted a comprehensive review of the literature on graph-based methods for facial affect analysis. These methods typically take an image or a sequence of images as input and produce an affect classification or regression as output. Based on their review, Liu et al. \cite{liu2022graph} proposed a pipeline for graph-based facial affect analysis, which includes (i) face preprocessing, (ii) graph-based affective representation, and (iii) graph-based relational reasoning. Preprocessing involves steps such as face detection and registration. Graph-based affective representation involves defining the structure of the graph, i.e., nodes and edges. The graph structure can be spatial or spatiotemporal depending on whether the input data is still images or videos. The graph structure can be at the landmark level, region level, or AU level, with nodes representing facial landmarks, facial regions of interest, or facial AUs, respectively. In the relational reasoning step, the edges, nodes, their interrelations, and temporal dynamics are analyzed through relational reasoning machine-learning or deep-learning models to make inferences regarding affect. Models used to analyze graph data include dynamic Bayesian networks, RNNs, CNNs, fully-connected neural networks, and non-temporal and temporal graph neural networks \cite{liu2022graph}.

Zhou et al. \cite{9191181} proposed a facial expression recognition method based on spatiotemporal landmark-level and region-level graphs. The intra-frame graph is formed by the connections among thirty-four facial landmarks located around the eyes, lips, and cheeks. The definition of these intra-frame connections was done manually. Inter-frame connections were established by linking each node in one frame to its corresponding node in the following frame. Two parallel ST-GCNs with analogous structures were trained; one on the nodes' $x$- and $y$-coordinates, and another on their histogram of orientation features. ST-GCNs' outputs were concatenated and processed by a fully connected network to identify facial expressions. This method's disadvantages include independent training of the two ST-GCNs rather than joint learning, and manual definition of nodes and edges.

Wei et al. \cite{wei2023geometric} presented a graph-based method for micro-expression recognition in video. The method included a dual-stream ST-GCN, focusing on the $x$ and $y$ coordinates of facial landmarks and the distance and angles between adjacent facial landmarks. An AU-specific loss function was incorporated into the neural network's training process in order to incorporate the association between AUs and micro-expressions. Their methodology employed three different sets of facial landmarks as graph nodes, comprising sets with 14, 31, and Dlib's \cite{king_dlib} 68 facial landmarks. Notably, the set with 68 landmarks was the only one to include landmarks along the jawline. The experiments demonstrated the best results for micro-expression recognition when 14 facial landmark sets were used.

Leong et al. \cite{fwa2022fine} introduced a method employing spatial-temporal graph attention networks to analyze facial landmarks and head poses, aimed at identifying academic emotions. The facial landmarks used were those around the eyebrow, eye, nose, and mouth excluding those that outline the face's outer shape, with the reasoning being they lack correlation with affective states. A notable limitation of this approach is its reliance on multiple deep neural networks for extracting features, i.e., facial landmarks and head pose. The method achieved lower academic emotion detection accuracy when compared to prior feature-based methods \cite{abedi2023affect}.

\vspace{-2mm}
\subsection{Discussion}
\vspace{-2mm}
\label{related_work_discussion}
As reviewed in this section, while some methods have used facial landmarks and ST-GCNs to recognize facial affect and expression in videos, there has been no exploration of the use of these methods to measure engagement. The fundamental differences between engagement and facial affect and expression make their measurement different. First, engagement is a multi-component state comprising behavioral, affective, and cognitive components. To illustrate, key indicators of behavioral engagement, such as being off-task or on-task, are determined by head pose, eye gaze, and blink rate. These indicators, and therefore engagement, cannot be effectively measured using methods designed solely for facial affect analysis. Second, engagement is not a constant state; it varies over time and should be measured at specific time resolutions where it remains stable and quantifiable. An ideal measurement duration for engagement is between ten and forty seconds, which is longer than the time resolution for facial affect analysis, which sometimes occurs at the frame level. Third, engagement measurement could involve recognizing an ordinal variable that indicates levels of engagement, as opposed to facial expression recognition, which identifies categorical variables without inherent order.

Existing engagement measurement approaches face limitations due to the necessity of employing multiple deep neural networks for the extraction and analysis of multi-modal features. Coupled with the significant differences between facial affect analysis and engagement measurement as outlined above, these limitations underscore a gap in the field. In response, we introduce a straightforward yet effective method for engagement measurement. This method is lightweight and fast, preserving privacy while also demonstrating improvements over current methodologies across two video-based engagement measurement datasets.

\vspace{-2mm}
\section{Method}
\label{sec:methodology}
\vspace{-2mm}
The input to the proposed method is a video sample of a student seated in front of a laptop or PC camera during a virtual learning session. The sequences of facial landmarks extracted from consecutive video frames through MediaPipe \cite{lugaresi2019mediapipe,grishchenko2020attention} are analyzed by ST-GCN \cite{yan2018spatial,10144079} to output the engagement level of the student in the video.

\vspace{-2mm}
\subsection{Graph-based Representation}
\label{sec:graph_representation}
\vspace{-2mm}
The MediaPipe deep learning solution \cite{lugaresi2019mediapipe}, a framework that is both real-time and cross-platform, is employed for the extraction of facial landmarks from video. Incorporated within MediaPipe, Attention Mesh \cite{grishchenko2020attention} is adept at detecting 468 3D facial landmarks throughout the face and an extra 10 landmarks for the iris. However, not all 478 landmarks are employed in the proposed engagement measurement method. Consistent with existing studies \cite{liu2022graph}, only 68 of the 3D facial landmarks, which match those identified by the Dlib framework \cite{king_dlib}, in addition to the 10 3D iris landmarks, making a total of 78 landmarks, are utilized.

The 3D facial landmarks encapsulate crucial spatial-temporal information pertinent to head pose \cite{malek2021head}, AUs \cite{jacob2021facial}, eye gaze \cite{grishchenko2020attention}, and affect \cite{liu2022graph,toisoul2021estimation}—key features identified in prior engagement measurement studies \cite{karimah2022automatic,dewan2019engagement,khan2022inconsistencies,ma2021automatic,copur2022engagement,abedi2023affect,abedi2023detecting,thomas2018predicting,gupta2016daisee,abedi2021improving}. Consequently, there is no necessity for extracting additional handcrafted features from the video frames.

The \(N\ \) 3D facial landmarks extracted from \( T \) consecutive video frames are utilized to construct a spatiotemporal graph \( G = (V, E) \). In this graph, the set of nodes \( V = \{v_{ti} | t = 1, \ldots, T, i = 1, \ldots, N\} \) encompasses all the facial landmarks in a sequence. To construct \( G \), first, the facial landmarks within one frame are connected with edges according to a connectivity structure based on Delaunay triangulation \cite{liu2019facial} which is consistent with true facial muscle distribution and uniform for different subjects \cite{liu2022graph}. Then each landmark will be connected to the same landmark in the consecutive frame.

\vspace{-2mm}
\subsection{Graph-based Reasoning}
\label{sec:graph_reasoning}
\vspace{-2mm}
Based on the spatiotemporal graph of facial landmarks \( G = (V, E) \) constructed above, an adjacency matrix \( A \) is defined as an \( N \times N \) matrix where the element at position \( (i, j) \) is set to 1 if there is an edge connecting the \( i^{th} \) and \( j^{th} \) landmarks, and set to 0 otherwise. An identity matrix \( I \), of the same dimensions as \( A \), is created to represent self-connections. A spatial-temporal graph, as the basic element of an ST-GCN layer, is implemented as follows \cite{yan2018spatial}.

\begin{equation}
    f_{\text{out}} = \left(\Lambda^{-\frac{1}{2}}((A + I) \odot M)\Lambda^{-\frac{1}{2}}f_{\text{in}}W_{\text{spatial}}\right)W_{\text{temporal}}
\end{equation}

\noindent
where \( \Lambda^{ii} = \sum_{j}(A^{ij} + I^{ij}) \). \(M\) is a learnable weight matrix that enables scaling the contributions of a node’s feature to its neighboring nodes \cite{yan2018spatial}. The input feature map, denoted \(f_{\text{in}}\), is the raw coordinates of facial landmarks for the first layer of ST-GCN, and it represents the outputs from previous layers in subsequent layers of ST-GCN. The dimensionality of \(f_{\text{in}}\) is \((C, N, T)\), where \(C\) is the number of channels, for example, 3 in the initial input to the network corresponding to the \(x\), \(y\), and \(z\) coordinates of the facial landmarks. In each layer of ST-GCN, initially, the spatial (intra-frame) convolution is applied to \(f_{\text{in}}\) based on the weight matrix \(W_{\text{spatial}}\), utilizing a standard 2D convolution with a kernel size of \( 1 \times 1\). Subsequently, the resulting tensor is multiplied by the normalized adjacency matrix \(\Lambda^{-\frac{1}{2}}((A + I) \odot M)\Lambda^{-\frac{1}{2}}\) across the spatial dimension. Afterward, the temporal (inter-frame) convolution, based on the weight matrix \(W_{\text{temporal}}\), is applied to the tensor output from the spatial convolution. This convolution is a standard 2D convolution with a kernel size of \( 1 \times \Gamma\), where \(\Gamma\) signifies the temporal kernel size.

Following an adequate number of ST-GCN layers, specifically three in this work, that perform spatial-temporal graph convolutions as outlined above, the resulting tensor undergoes 2D average pooling. The final output of the network is generated by a final 2D convolution. This convolution employs a kernel size of \( 1 \times 1 \) and features an output channel dimensionality equal to the number of classes \(K\), i.e., the number of engagement levels to be measured. An explanation of the detailed architecture of the ST-GCNs for specific datasets can be found in subsection \ref{sec:cexperimental_setting}.

\vspace{-2mm}
\subsection{Ordinal Engagement classification through Transfer Learning}
\label{sec:ordinal_classification}
\vspace{-2mm}
The model described above, with a final layer comprising \( K \) output channels, tackles the engagement measurement problem as a categorical \( K \)-class classification problem without taking into account the ordinal nature of the engagement variable \cite{yannakakis2018ordinal,whitehill2014faces,khan2022inconsistencies}. The model could harness the ordinality of the engagement variable to enhance its inferences. Drawing inspiration from \cite{frank2001simple,abedi2023affect}, a novel ordinal learning framework based on transfer learning is introduced as follows.

\noindent
\textbf{Training phase–}
The original \(K\)-level ordinal labels, \(y = 0, 1, \ldots, K - 1\), in the training set are converted into \(K - 1\) binary labels \(y_i\) as follows: if \(y > i\), then \(y_i = 1\); otherwise, \(y_i = 0\), for \(i = 0, 1, \ldots, K - 2\). Subsequently, \(K - 1\) binary classifiers are trained with the training set and the \(K - 1\) binary label sets described above. The training of binary classifiers is based on transfer learning, which proceeds as follows: Initially, a network is trained on the dataset with the original \(K\)-class labels, employing a regular final layer with \(K\) output channels. After training, the ST-GCN layers of this network are frozen, and the final layer is removed. To this frozen network, \(K - 1\) separate untrained 2D convolution layers with a single output channel each are added, resulting in \(K - 1\) new networks. Each of these networks consists of a frozen sequence of ST-GCN layers followed by an untrained 2D convolution layer. These \(K - 1\) new networks are then trained on the entire dataset using the \(K - 1\) binary label sets described above. During this phase, only the final 2D convolution layers are subjected to training.

\noindent
\textbf{Inference phase–}
For the ordinal classification of a test sample, the sample is initially input into the pre-trained (and frozen) sequence of ST-GCN layers, followed by a 2D average pooling layer. The tensor obtained from this process is then input into \(K - 1\) pre-trained final 2D convolution layers, each yielding a probability estimate for the test sample being in the binary class \(y_t = y_i\), where \(i = 0, 1, \ldots, K - 2\). Subsequently, these \(K - 1\) binary probability estimates are transformed into a single multi-class probability of the sample belonging to class \(y = 0, 1, \ldots, K - 1\), as follows \cite{frank2001simple}.

\begin{equation}
    p(y_t = k) = 
    \begin{cases} 
        1 - p(y_t \geq 0), & \text{if } k = 0, \\
        p(y_t > k - 1) - p(y_t \geq k), & \text{if } 0 < k < K - 1, \\
        p(y_t > K - 2), & \text{if } k = K - 1.
    \end{cases}
\end{equation}

Despite the increased training time for the ordinal model within the aforementioned ordinal learning framework, the final count of parameters in the ordinal model remains nearly identical to that of the original non-ordinal model.

\vspace{-2mm}
\section{Experiments}
\label{sec:experiments}
\vspace{-2mm}
This section evaluates the performance of the proposed method relative to existing methods in video-based engagement measurement. It reports and discusses the results of multi-class and binary classification of engagement across two datasets. Based on the engagement measurement problem at hand, several evaluation metrics are employed. In the context of multi-class engagement level classification, metrics such as accuracy and confusion matrix are reported. For binary engagement classification, accuracy, the Area Under the Curve of the Receiver Operating Characteristic (AUC-ROC), and the Area Under the Curve of the Precision and Recall curve (AUC-PR) are utilized. In addition, the number of parameters of the models used, memory consumption, and inference time of the proposed method are compared to those of previous methods.

\vspace{-2mm}
\subsection{Datasets}
\label{datasets}
\vspace{-2mm}
Experiments on two large video-based engagement measurement datasets were conducted, each presenting unique challenges that further enabled the validation of the proposed method.

\noindent
\textbf{EngageNet:}
The EngageNet dataset \cite{engagenet}, recognized as the largest dataset for student engagement measurement, includes video recordings of 127 subjects participating in virtual learning sessions. Each video sample has a duration of 10 seconds, with a frame rate of 30 fps and a resolution of \(1280 \times 720\) pixels. The subjects' video recordings were annotated as four ordinal levels of engagement: Not-Engaged, Barely-Engaged, Engaged, and Highly-Engaged. The dataset was divided into 7983, 1071, and 2257 samples for training, validation, and testing, respectively, using a subject-independent data split approach \cite{engagenet}. However, only the training and validation sets were made available by the dataset creators and were utilized for the training and validation of predictive models in the experiments presented in this paper. The distribution of samples in the four aforementioned classes of engagement in the training and validation sets are 1550, 1035, 1658, and 3740, and 132, 97, 273, and 569, respectively.

\noindent
\textbf{Online SE:}
The Online SE dataset \cite{thomas2022automatic}, comprises videos of six students participating in online courses via the Zoom platform. These recordings span 10 seconds each, with a frame rate of 24 fps and a resolution of $220 \times 155$ pixels. The videos were annotated as either Not-Engaged or Engaged. The dataset was segmented into 3190, 1660, and 1290 samples for training, validation, and testing, respectively. The distribution of samples in the two aforementioned classes of engagement in the training, validation, and test sets is 570 and 2620, 580 and 1080, and 570 and 720, respectively.

\vspace{-2mm}
\subsection{Experimental Setting}
\label{sec:cexperimental_setting}
\vspace{-2mm}
The sole information extracted from video frames is facial landmarks, which are analyzed by ST-GCN to determine the engagement level of the student in the video.
Drawing inspiration from the pioneering works on body-joints-based action analysis \cite{yan2018spatial,zheng2023skeleton}, the proposed ST-GCN for facial-landmarks-based engagement measurement is structured as follows. The input facial landmarks are first processed through a batch normalization layer, followed by three consecutive ST-GCN layers with 64, 128, and 256 output channels, respectively. Residual connections are incorporated in the last two ST-GCN layers. A dropout rate of 0.1 is applied to each ST-GCN layer. The temporal kernel size in the ST-GCN layers is selected to be \(9\). Subsequent to the ST-GCN layers, an average pooling layer is utilized, and its resulting tensor is directed into a 2D convolutional layer with 256 input channels and a number of output channels corresponding to the number of classes. A Softmax activation function then computes the probability estimates. In cases where engagement measurement is framed as a binary classification task, the terminal 2D convolution layer is configured with a single output channel, substituting Softmax with a Sigmoid function. The Sigmoid function is also employed for the individual binary classifiers within the ordinal learning framework detailed in subsection \ref{sec:ordinal_classification}. The models are trained using the Adam optimizer with mini-batches of size \(16\) and an initial learning rate of \(0.001\) for \(300\) epochs. The learning rate is decayed by a factor of \(0.1\) every \(100\) epochs.

\begin{table}[h!]
\caption{Classification accuracy of engagement levels on the validation set of the EngageNet dataset \cite{engagenet}: comparison of state-of-the-art end-to-end methods and feature-based methods with various feature sets and classification models against two configurations of the proposed method - facial landmarks analyzed by ST-GCN and ordinal ST-GCN. Bolded values denote the best results.}
\label{tab:engagenet_comparison}
\centering
\renewcommand{\arraystretch}{0.7} 
\footnotesize
\begin{tabular}{p{.1\linewidth}p{.45\linewidth}p{.3\linewidth}p{.1\linewidth}}
\hline
Ref. & Features & Model & Accuracy\\
\hline
\hline

\cite{abedi2021improving} & End to End Model & ResNet + TCN & 0.5472\\
\hline
\cite{selim2022students} & End to End Model & EfficientNet + LSTM & 0.5757\\
\hline
\cite{selim2022students} & End to End Model & EfficientNet + Bi-LSTM & 0.5894\\
\hline
\hline

\cite{engagenet} & Gaze & LSTM & 0.6125\\
\hline
\cite{engagenet} & Head Pose & LSTM & 0.6760\\
\hline
\cite{engagenet} & AU & LSTM & 0.6303\\
\hline
\cite{engagenet} & Gaze + Head Pose & LSTM & 0.6769\\
\hline
\cite{engagenet} & Gaze + Head Pose + AU & LSTM & 0.6704\\
\hline
\hline

\cite{engagenet} & Gaze & CNN-LSTM & 0.6060\\
\hline
\cite{engagenet} & Head Pose & CNN-LSTM & 0.6732\\
\hline
\cite{engagenet} & AU & CNN-LSTM & 0.6172\\
\hline
\cite{engagenet} & Gaze + AU & CNN-LSTM & 0.6275\\
\hline
\cite{engagenet} & Head Pose + AU & CNN-LSTM & 0.6751\\
\hline
\cite{engagenet} & Gaze + Head Pose + AU & CNN-LSTM & 0.6751\\
\hline
\hline

\cite{engagenet} & Gaze & TCN & 0.6256\\
\hline
\cite{engagenet} & Head Pose & TCN & 0.6611\\
\hline
\cite{engagenet} & AU & TCN & 0.6293\\
\hline
\cite{engagenet} & Gaze + Head Pose + AU & TCN & 0.6779\\
\hline
\hline

\cite{engagenet} & Gaze & Transformer & 0.5545\\
\hline
\cite{engagenet} & Gaze + Head Pose  & Transformer & 0.6445\\
\hline
\cite{engagenet} & Gaze + Head Pose + AU & Transformer & 0.6910\\
\hline
\cite{engagenet} & Gaze + Head Pose + AU + MARLIN & Transformer & 0.6849\\
\hline
\hline

\cite{vedernikov2024tcct} & Gaze & TCCT-Net & 0.6433\\
\hline
\cite{vedernikov2024tcct} & Head Pose & TCCT-Net & 0.6891\\
\hline
\cite{vedernikov2024tcct} & AU & TCCT-Net & 0.6629\\
\hline
\cite{vedernikov2024tcct} & Gaze + Head Pose  & TCCT-Net & 0.6564\\
\hline
\cite{vedernikov2024tcct} & Gaze + Head Pose + AU & TCCT-Net & 0.6713\\
\hline

\hline
\hline

Ours & Facial Landmarks & ST-GCN & 0.6937\\
\hline
Ours & Facial Landmarks & Ordinal ST-GCN & \textbf{0.7124}\\
\hline

\end{tabular}
\end{table}

\vspace{-2mm}
\subsection{Experimental Results}
\label{sec:experimental_results}
\vspace{-2mm}
\subsubsection{Comparison to Previous Methods}
\label{sec:experimental_results_comparison}
Table \ref{tab:engagenet_comparison} presents the comparative results of two settings of the proposed method, a regular non-ordinal classifier and an ordinal classifier, with previous methods on the validation set of the EngageNet dataset \cite{engagenet}. The engagement measurement in EngageNet \cite{engagenet} is a four-class classification problem and the accuracy is reported as the evaluation metric. The previous methods in Table \ref{tab:engagenet_comparison} include state-of-the-art end-to-end methods, including the combination of ResNet-50 with TCN \cite{abedi2021improving} and the combination of EfficientNet B7 with LSTM and bidirectional LSTM \cite{selim2022students}, followed by state-of-the-art feature-based methods. The previous feature-based methods in Table \ref{tab:engagenet_comparison} used different combinations of OpenFace's eye gaze, head pose, and AU features \cite{baltrusaitis2018openface} along with MARLIN's facial embedding features \cite{cai2023marlin}. These features were classified by LSTM, CNN-LSTM, TCN, and Transformer. Refer to \cite{engagenet} for more details. The results of the feature-based method proposed by Vedernikov et al. \cite{vedernikov2024tcct} are also reported, where the aforementioned features are classified using a Transformer-based neural network called the Tensor-Convolution and Convolution-Transformer Network (TCCT-Net).

Despite the abundant data samples in the EngageNet dataset \cite{engagenet} available for training complex neural networks such as ResNet + TCN \cite{abedi2021improving} and EfficientNet B7 + bidirectional LSTM \cite{selim2022students}, their performance is inferior to that of feature-based methods. This highlights the necessity of extracting hand-crafted features or facial landmarks from videos and building classifiers on top of them.

In the single feature configurations of previous methods in Table \ref{tab:engagenet_comparison}, head pose achieves better results compared to AUs, which is better than eye gaze. While head pose and eye gaze are indicators of behavioral engagement \cite{abedi2023affect,ocumpaugh2015baker}, AUs, which are associated with facial expressions and affect \cite{toisoul2021estimation}, are indicators of affective engagement. Combining these three features is always beneficial since engagement is a multi-component variable that can be measured by affective and behavioral indicators when the only available data is video \cite{abedi2023affect}. Among classification models, utilizing more advanced models improves accuracy; the Transformer is better than TCN, which is better than CNN-LSTM and LSTM; however, it comes at the cost of increased computational complexity. For single features, the Transfomer-based TCCT-Net \cite{vedernikov2024tcct} outperforms other classifiers; however, for multiple feature sets, the vanilla Transformer \cite{engagenet} outperforms the others.

The proposed ST-GCN in Table \ref{tab:engagenet_comparison}, which relies solely on facial landmarks without requiring raw facial videos or multiple hand-crafted features, outperforms previous methods. Moreover, making the proposed method ordinal further improves the state-of-the-art by 3.1\% compared to eye gaze, head pose, AUs, and MARLIN features \cite{cai2023marlin} with the Transformer \cite{engagenet}. Tables \ref{tab:confusion_matrix_a} and \ref{tab:confusion_matrix_b} depict the confusion matrices of the ordinal and non-ordinal configurations of the proposed method in Table \ref{tab:engagenet_comparison}. As shown, incorporating ordinality significantly increases the number of correctly classified samples in the first three classes and results in a 2.7\% improvement in accuracy compared to its non-ordinal counterpart.

\vspace{-10pt} 
\begin{table}[ht]
\setlength{\abovecaptionskip}{0pt} 
\setlength{\belowcaptionskip}{0pt} 
\renewcommand{\arraystretch}{0.8} 
\centering
\caption{Confusion matrices of the proposed method with (a) non-ordinal and (b) ordinal ST-GCN on the validation set of the EngageNet dataset \cite{engagenet}.}
\label{tab:confusion_matrix}

\begin{subtable}{.45\linewidth}
\centering
\caption{Non-ordinal ST-GCN}
\label{tab:confusion_matrix_a}
\setlength{\tabcolsep}{6pt} 
\footnotesize
\begin{tabular}{@{}ccccc@{}}
\toprule
\textbf{Class} & \textbf{1} & \textbf{2} & \textbf{3} & \textbf{4} \\ \midrule
\textbf{1} & 99 & 14 & 13 & 6 \\
\textbf{2} & 10 & 29 & 33 & 25 \\
\textbf{3} & 9 & 19 & 100 & 145 \\
\textbf{4} & 5 & 4 & 45 & 515 \\ \bottomrule
\end{tabular}
\end{subtable}%
\hspace{0.05\linewidth} 
\begin{subtable}{.45\linewidth}
\centering
\caption{Ordinal ST-GCN}
\label{tab:confusion_matrix_b}
\setlength{\tabcolsep}{6pt} 
\footnotesize
\begin{tabular}{@{}ccccc@{}}
\toprule
\textbf{Class} & \textbf{1} & \textbf{2} & \textbf{3} & \textbf{4} \\ \midrule
\textbf{1} & 104 & 12 & 10 & 6 \\
\textbf{2} & 15 & 36 & 24 & 22 \\
\textbf{3} & 10 & 26 & 112 & 125 \\
\textbf{4} & 9 & 5 & 44 & 511 \\ \bottomrule
\end{tabular}
\end{subtable}

\end{table}
\vspace{-10pt} 

\vspace{-10pt} 
\begin{table}[ht]
\caption{Classification accuracy of engagement levels on the validation set of the EngageNet dataset \cite{engagenet} for different variants of the proposed method.}
\label{tab:variants}
\centering
\renewcommand{\arraystretch}{0.8} 
\footnotesize
\begin{tabular}{p{.75\linewidth}p{.1\linewidth}}
\hline
Variant & Accuracy \\ \hline \hline
ST-GCN is replaced with LSTM & 0.6847 \\ \hline
ST-GCN is replaced with TCN & 0.6813 \\ \hline
Only \(x\) and \(y\) coordinates of joints are used & 0.6655 \\ \hline
A temporal kernel of 3 is used & 0.6748 \\ \hline
A temporal kernel of 15 is used & 0.6841 \\ \hline
Every 2 frames are used & 0.6813 \\ \hline
Every 4 frames are used & 0.6907 \\ \hline
Every 8 frames are used & 0.6907 \\ \hline
Every 16 frames are used & 0.6841 \\ \hline
Hand landmarks are added & 0.6956 \\ \hline
\end{tabular}
\end{table}
\vspace{-10pt} 

The feature extraction step in \cite{vedernikov2024tcct} and \cite{engagenet} involves running multiple deep-learning models in OpenFace \cite{baltrusaitis2018openface} to capture eye gaze, head pose, and AUs, as well as another complex network for MARLIN feature embeddings \cite{cai2023marlin}. While these feature extraction networks are complex for real-time use, the proposed method relies on facial landmarks extracted using the real-time MediaPipe \cite{lugaresi2019mediapipe, grishchenko2020attention}. Considering only the classification models, the number of parameters in EfficientNet B7 + LSTM \cite{selim2022students}, ResNet + TCN \cite{abedi2021improving}, Transformer \cite{engagenet}, the proposed non-ordinal ST-GCN, and ordinal ST-GCN are 82,681,812, 24,639,236, 1,063,108, 861,688, and 861,431, respectively. The memory consumption for EfficientNet B7 + LSTM \cite{selim2022students}, ResNet + TCN \cite{abedi2021improving}, Transformer \cite{engagenet}, and the proposed ordinal ST-GCN are 2268.78, 790.35, 178.00, and 180.96 megabytes, respectively. The inference time for classifying a data sample using EfficientNet B7 + LSTM \cite{selim2022students}, ResNet + TCN \cite{abedi2021improving}, Transformer \cite{engagenet}, and the proposed ordinal ST-GCN are 348, 51, 10, and 0.8 milliseconds, respectively. This indicates the efficiency of the proposed method, which, while being lightweight and fast, also improves the state-of-the-art.

\vspace{-2mm}
\subsubsection{Variants of the Proposed Method}
\label{sec:experimental_results_variants}
Table~\ref{tab:variants} displays the results of different variants of the proposed method on the validation set of the EngageNet \cite{engagenet} dataset. In the first two variants, the \( x \), \( y \), and \( z \) coordinates of facial landmarks are converted into multivariate time series and analyzed by an LSTM and TCN. The LSTM includes four unidirectional layers with 256 neurons in hidden units and is followed by a \( 256 \times 4 \) fully connected layer. The parameters of the TCN are as follows: the number of layers, number of filters, kernel size, and dropout rate are 8, 64, 8, and 0.05, respectively. While their results are acceptable and better than most of the earlier methods in Table~\ref{tab:engagenet_comparison}, they cannot outperform ST-GCN, the last two rows of Table~\ref{tab:engagenet_comparison}. The fact that the accuracy of the LSTM and TCN in Table~\ref{tab:variants} is higher than those in Table~\ref{tab:engagenet_comparison} signifies the efficiency of facial landmarks for engagement measurement. In the third variant, the \( z \) coordinates of facial landmarks are disregarded, and the decrease in the accuracy of the non-ordinal ST-GCN indicates the importance of the \( z \) coordinates for engagement measurement. Temporal kernel sizes other than 9 in the fourth and fifth variants have a negative impact on the results of the non-ordinal ST-GCN. When engagement measurement is performed using every 2, 4, 8, and 16 frames, instead of every frame, there is a slight decrease in the accuracy of the non-ordinal ST-GCN. However, this is a trade-off between accuracy and computation since reducing the frame rate corresponds to a reduction in computation. The last row of Table~\ref{tab:variants} shows the results of the ordinal ST-GCN when 21 hand landmarks extracted using MediaPipe \cite{lugaresi2019mediapipe} were added to the facial landmarks. The lower accuracy compared to using only facial landmarks is due to two factors. Firstly, there is 80\% missingness in the hand landmarks due to the absence of hands in the videos. Secondly, it indicates that facial landmarks alone are sufficient for engagement measurement, capturing both behavioral and emotional indicators of engagement.

\vspace{-2mm}
\subsubsection{Results on the Online SE Dataset}
\label{sec:experimental_results_online}
Table~\ref{tab:se_comparison} presents the results of the proposed method in comparison to previous methods on the test set of the Online SE dataset \cite{thomas2022automatic}. The earlier methods listed in Table~\ref{tab:se_comparison} are feature-based, extracting affective and behavioral features from video frames and performing binary classification of the features using TCN in \cite{thomas2018predicting}, TCN in \cite{abedi2023affect}, LSTM with attention in \cite{chen2019faceengage}, and BoW in \cite{abedi2023bag}. Given that engagement measurement in the Online SE dataset \cite{thomas2022automatic} is posed as a binary classification problem, implementing the ordinal version of the proposed method is not required. The proposed method, employing facial landmarks with ST-GCN, attains the highest accuracy and AUC PR.

\vspace{-10pt} 
\begin{table}[h]
\caption{Binary classification accuracy, AUC ROC, and AUC PR of engagement on the test set of the Online SE dataset \cite{thomas2022automatic}: comparison of previous methods against the proposed method. Bolded values denote the best results.}
\label{tab:se_comparison}
\centering
\renewcommand{\arraystretch}{0.8} 
\footnotesize
\begin{tabular}{p{.3\linewidth}p{.15\linewidth}p{.15\linewidth}p{.15\linewidth}}
\hline
Method & Accuracy & AUC ROC & AUC PR \\ \hline \hline
\cite{thomas2018predicting} & 0.7803 & 0.8764 & 0.8008 \\ \hline
\cite{abedi2023affect} & 0.7637 & 0.8710 & 0.7980 \\ \hline
\cite{chen2019faceengage} & 0.7475 & 0.8890 & 0.7973 \\ \hline
\cite{abedi2023bag} & 0.8191 & \textbf{0.8926} & 0.9018 \\ \hline
Proposed & \textbf{0.8315} & 0.8806 & \textbf{0.9131} \\ \hline
\end{tabular}
\end{table}
\vspace{-10pt} 

\vspace{-2mm}
\subsubsection{Interpretation of Results}
\label{sec:experimental_results_interpretation}
\begin{figure}[h!]
    \centering
    \includegraphics[width=.95\textwidth]{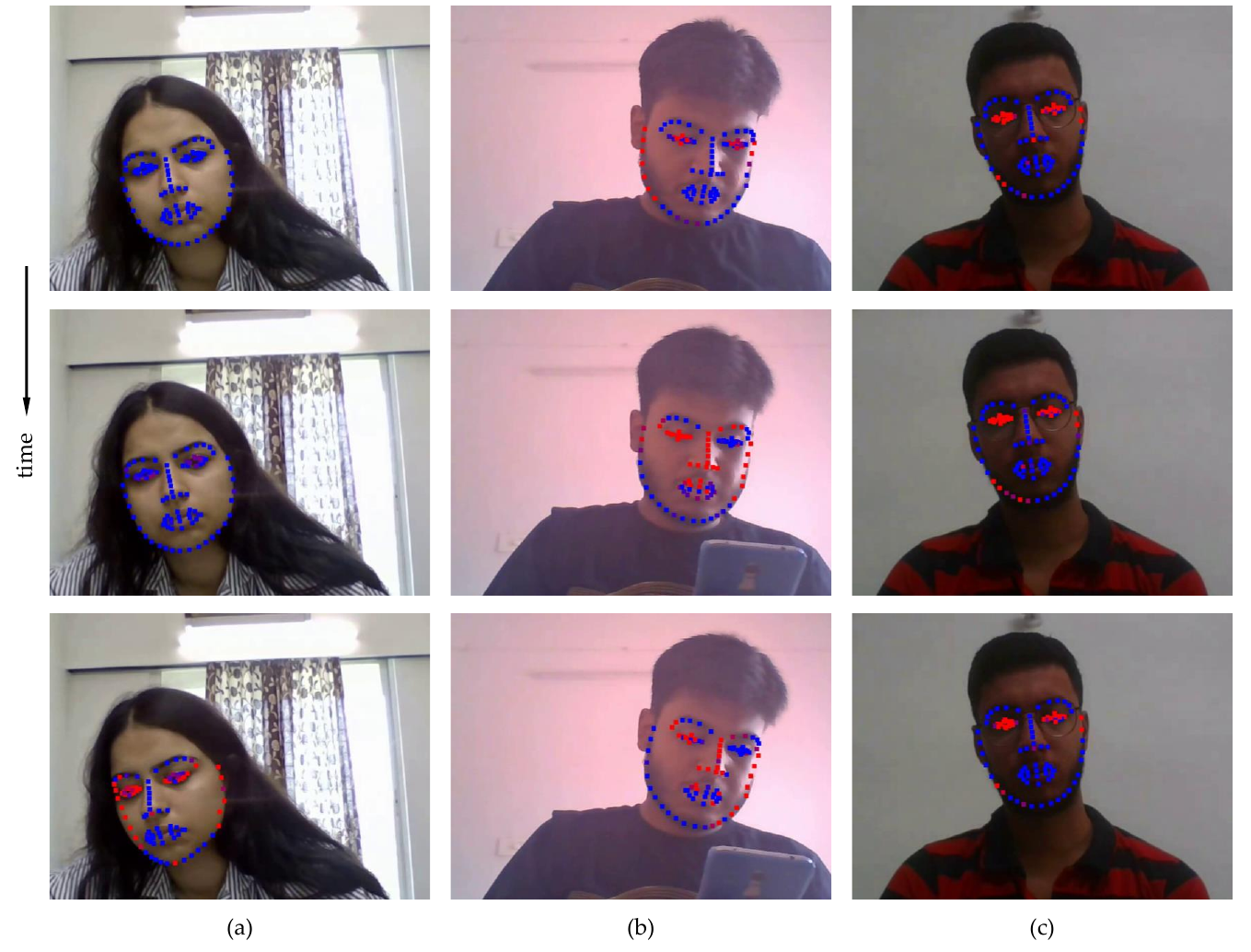}
    \caption{Interpretation of engagement measurements taken using the proposed method by applying Gradient-weighted Class Activation Mapping (Grad-CAM) to the Spatial-Temporal Graph Convolutional Network (ST-GCN). Please refer to the last paragraph of subsection \ref{sec:experimental_results_interpretation} for further details.}
    \label{fig:gradcam}
\end{figure}

Fig. \ref{fig:gradcam} displays the interpretation of engagement measurements taken using the proposed method by applying Grad-CAM \cite{zheng2023skeleton} to the ordinal ST-GCN trained on the training set of the EngageNet dataset. For visualization purposes, three exemplary frames (out of 300) from the beginning, middle, and end of three data samples annotated as Not-Engaged in the validation set of the EngageNet dataset are shown in Fig. \ref{fig:gradcam} (a)-(c). The facial landmarks extracted through MediaPipe are overlaid on the frames, where the color map of the facial landmarks, from blue to red, depicts the class activation map values of the last ST-GCN layer in the trained ST-GCN. The facial landmarks associated with the target class of Not-Engaged at certain frames are colored towards red. In Fig. \ref{fig:gradcam} (a), during the beginning and middle of the video, the first two exemplary frames show the student engaged, with lower class activation map values. At the end of the video, when the student is not looking at the camera (computer screen) and is looking elsewhere, landmarks on the iris, eye, and jawline show higher values, resulting in the model classifying the sample as Not-Engaged. In Fig. \ref{fig:gradcam} (b), the student is not paying attention and is playing with their phone. Jawline, eye, iris, and eyebrow facial landmarks with red colors have higher class activation map values, resulting in the model classifying the sample as Not-Engaged. In Fig. \ref{fig:gradcam} (c), throughout the entire video sample, the head pose of the student is normal and perpendicular to the camera. However, the eyes are almost closed, indicating sleepiness and low arousal. This is detected by the higher class activation map values on the eye and iris landmarks, corresponding to the intensity of AU number 45, which indicates how closed the eyes are. This resulted in the model classifying the sample as Not-Engaged.

\vspace{-5pt} 

\vspace{-2mm}
\section{Discussion}
\label{sec:conclusion}
\vspace{-2mm}
Our research led to the development of a novel deep-learning framework for student engagement measurement. In the proposed framework, sequences of facial landmarks are extracted from consecutive video frames and analyzed by ordinal ST-GCNs to make inferences regarding the engagement level of the student in the video. The successful application of our model to the EngageNet \cite{engagenet} and Online SE \cite{thomas2022automatic} datasets not only confirms its efficacy but also establishes a new standard in engagement level classification accuracy, outperforming previous methods. As the sole input information to the developed engagement measurement models, the 3D facial landmarks contain information about head pose \cite{malek2021head}, AUs \cite{jacob2021facial}, eye gaze \cite{grishchenko2020attention}, and affect \cite{liu2022graph,toisoul2021estimation}, which are the key indicators of behavioral and affective engagement. The relatively lightweight and fast ST-GCN and its integration with real-time MediaPipe \cite{lugaresi2019mediapipe} make the proposed framework capable of being deployed on virtual learning platforms and measuring engagement in real-time. The proposed method is privacy-preserving and does not require access to personally identifiable raw video data for engagement measurement. In a real-world deployment, the cross-platform MediaPipe solution \cite{lugaresi2019mediapipe}, running on a web, mobile, or desktop application, extracts facial landmarks from video data on users' local devices. These non-identifiable facial landmarks are then transferred to a cloud, where they are analyzed by ST-GCNs to measure engagement. The interpretability feature of the proposed method, enabled through Grad-CAM, facilitates understanding which facial landmarks, corresponding to behavioral and affective indicators of engagement, contribute to certain levels of engagement. It also helps identify the specific timestamps at which these contributions occur. This provides instructors with additional information to take necessary actions and promote student engagement. A limitation of the proposed method is its reliance on the quality of facial landmarks detected by MediaPipe. In the context of engagement measurement in virtual learning sessions, an occluded or absent face, and consequently non-detected facial landmarks, correspond to lower levels of engagement or disengagement. Our developed ST-GCN was able to correctly classify most samples with occluded or absent faces, i.e., no facial landmarks, as Not-Engaged. To improve the performance of the proposed method, the following direction could be investigated: analyzing facial landmarks with more advanced ST-GCNs, which are equipped with attention mechanisms and trained through contrastive learning techniques and applying augmentation techniques to video data before facial landmark extraction \cite{abedi2023cross} or to facial landmark data to improve the generalizability of ST-GCNs.

\vspace{\baselineskip}
\noindent
\textbf{Acknowledgment } The authors express their sincere thanks to the Multimodal Perception Lab at the International Institute of Information Technology, Bangalore, India, for their generosity in providing the Online SE dataset, which was instrumental in the execution of our experiments.

\noindent
This research was funded by the Natural Sciences and Engineering Research Council of Canada.


\bibliographystyle{IEEEtran}
\bibliography{references}

\end{document}